\documentclass[11pt]{article}

\usepackage[final]{acl}

\usepackage{times}
\usepackage{latexsym}

\usepackage[T1]{fontenc}

\usepackage[utf8]{inputenc}

\usepackage{microtype}

\usepackage{inconsolata}

\usepackage{graphicx}

\usepackage{amsmath}
\usepackage{caption}
\usepackage{enumitem}
    \setlist{
      topsep=3pt,   
      itemsep=-3pt,  
    }
\usepackage{subcaption}
\usepackage{booktabs}
\usepackage{listings}
\usepackage[most]{tcolorbox}
\usepackage{ulem}

\title{Rethinking Meeting Effectiveness: A Benchmark and Framework for Temporal Fine-grained Automatic Meeting Effectiveness Evaluation}

\author{Yihang Li, Chenhui Chu \\
  Kyoto University \\
  \texttt{liyh@nlp.ist.i.kyoto-u.ac.jp}, \texttt{chu@i.kyoto-u.ac.jp}
}

\begin{document}
\maketitle

\begin{abstract}
Evaluating meeting effectiveness is crucial for improving organizational productivity. Current approaches rely on post-hoc surveys that yield a single coarse-grained score for an entire meeting. The reliance on manual assessment is inherently limited in scalability, cost, and reproducibility. Moreover, a single score fails to capture the dynamic nature of collaborative discussions. We propose a new paradigm for evaluating meeting effectiveness centered on novel criteria and temporal fine-grained approach. We define effectiveness as the rate of objective achievement over time and assess it for individual topical segments within a meeting. To support this task, we introduce the AMI Meeting Effectiveness (AMI-ME) dataset, a new meta-evaluation dataset containing 2,459 human-annotated segments from 130 AMI Corpus meetings. We also develop an automatic effectiveness evaluation framework that uses a Large Language Model (LLM) as a judge to score each segment's effectiveness relative to the overall meeting objectives. Through substantial experiments, we establish a comprehensive benchmark for this new task and evaluate the framework's generalizability across distinct meeting types, ranging from business scenarios to unstructured discussions. Furthermore, we benchmark end-to-end performance starting from raw speech to measure the capabilities of a complete system. Our results validate the framework's effectiveness and provide strong baselines to facilitate future research in meeting analysis and multi-party dialogue. 
The AMI-ME dataset and the Automatic Evaluation Framework are available at \url{https://github.com/ku-nlp/AMI-ME}.

\end{abstract}

\begin{figure}[t]
  \centering
  \includegraphics[width=0.9\linewidth]{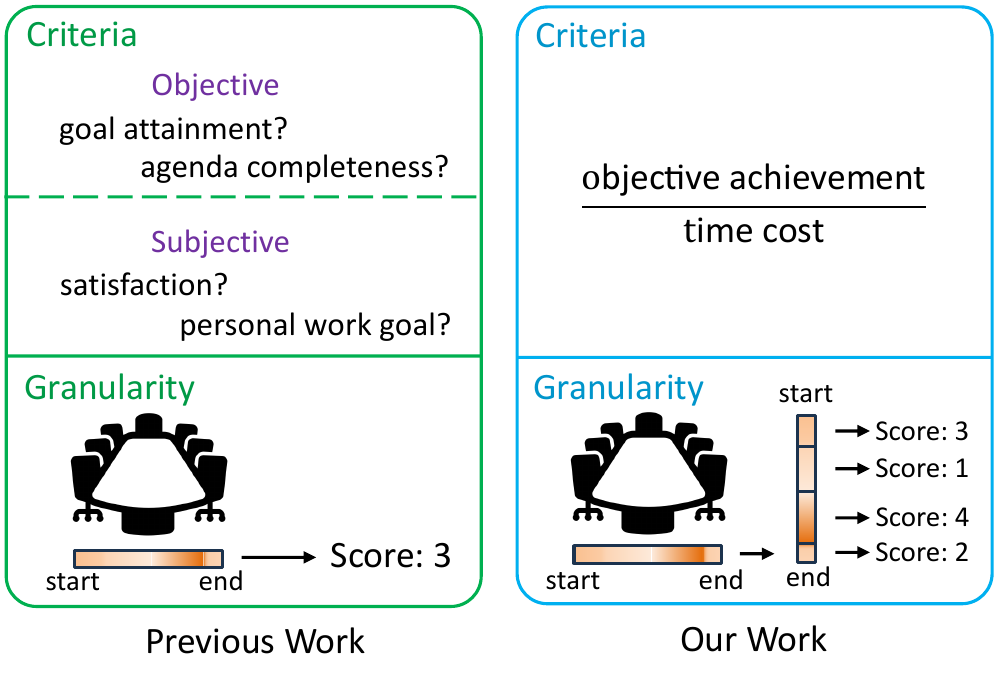}
  \caption{The paradigm of meeting effectiveness evaluation.}
  \label{fig:abstract}
\end{figure}

\vspace{-1mm}
\section{Introduction}

Meetings are a cornerstone of organizational collaboration. However, as their frequency and duration steadily increase \cite{52, tobia1990making}, they are often cited as a primary source of inefficiency and dissatisfaction, consuming vast amounts of time and resources \cite{Meeting_analysis, Not_Another_Meeting, Allen2022TheKF}. On the other hand, in the nascent field of multi-party dialogue\cite{castillo-lopez-etal-2025-survey}, meetings are almost the most important scenario, and the evaluation of meeting effectiveness constitutes the most crucial perspective. The challenge of improving meeting productivity hinges on the ability to measure effectiveness first. However, there is no common consensus on how to evaluate meeting effectiveness \cite{Meeting_Effectiveness_and_Inclusiveness}.

Current evaluation approaches typically rely on post-hoc participant surveys, which provide a single holistic score for an entire meeting \cite{Meeting_Effectiveness_and_Inclusiveness, nixon1992impact, WebRTC, 10.1145/3449247, telepresence, ComFeel, Leach2009Perceived}. 
These coarse-grained evaluations capture only a single retrospective impression of a meeting, overlooking the dynamic nature of collaborative sessions and failing to distinguish between productive and unproductive periods. Furthermore, this reliance on manual evaluation is costly, difficult to scale, and lacks reproducibility. Consequently, research in this area is often constrained by data scarcity, as collecting and annotating meeting data is time-consuming and raises significant privacy concerns.

The recent emergence of the ``LLM-as-a-Judge'' paradigm \cite{initial_llm_judge, gu2025surveyllmasajudge} offers a promising solution for automatic evaluation. This approach has gained significant traction for combining the scalability of automatic methods with the context-sensitivity of human experts. This breakthrough enables the development of an automatic framework for evaluating meeting effectiveness, which in turn is a critical prerequisite for creating intelligent multi-party dialogue agents capable of proactive intervention.

To this end, we introduce a new paradigm for meeting effectiveness evaluation centered on two core principles, as shown in Figure \ref{fig:abstract}. First, we propose an objective and universally applicable evaluation criterion that defines effectiveness as \textit{the rate of objective achievement over time}. This moves beyond subjective feelings to a more quantifiable measure of output versus input. Second, we propose a temporal fine-grained evaluation method. Instead of assigning a single score to an entire meeting, we divide the meeting into coherent topical segments and evaluate the effectiveness of each one. This approach offers three key advantages: (1) it enables detailed analysis of meeting dynamics; (2) it improves the reliability of human annotation by focusing on shorter and more manageable segments; and (3) it significantly amplifies the amount of data available for computational modeling, thereby helping to overcome the data scarcity bottleneck.

Based on this paradigm, we constructed a new meta-evaluation dataset for meeting effectiveness, which we call AMI-ME. We built upon the AMI Corpus \cite{AMI}, whose simulated scenarios have relatively low technical complexity but high realism, making the content accessible to annotators. We first refined the original coarse and discontinuous topic segmentations into continuous fine-grained units using a reference-based segmentation method. Subsequently, we conducted a comprehensive human annotation effort to score the effectiveness of each segment, implementing various rigorous quality control measures. The resulting dataset contains 2,459 annotated segments from 130 meetings, with fine-grained topic segmentations and the corresponding segment-level effectiveness scores.

Furthermore, we propose an LLM-based framework for the automatic meeting effectiveness evaluation. Our framework performs topic segmentation on the meeting transcript and then evaluates each segment's effectiveness based on its contribution to the classified overall meeting objectives. Through substantial experiments, we benchmark various LLMs, examine the framework's performance on distinct meeting types, and measure the performance of a full end-to-end system starting from raw speech. Our results validate the framework's effectiveness and provide strong baselines to facilitate future research into meeting analysis and multi-party dialogue agents.

In summary, our contributions are fourfold:
\begin{itemize}
    \item We propose a novel methodology for temporal fine-grained meeting effectiveness evaluation, based on the criteria of objective achievement over time.
    \item We construct the AMI-ME dataset for meeting effectiveness meta-evaluation, including fine-grained continuous topic segmentation and corresponding segment-level effectiveness scores.
    \item We propose an LLM-based framework to automatically evaluate meeting effectiveness at a fine-grained level.
    \item We conduct substantial experiments to establish strong benchmark results on the AMI-ME dataset.
\end{itemize}

\section{Related Work}
\label{subsec:mee}
\textbf{Meeting Effectiveness Evaluation.} 
Despite significant research in meeting science, there is no consensus on how to measure it. Prior work has predominantly relied on post-hoc participant surveys to derive a single holistic score for an entire meeting, with evaluation criteria varying widely across studies.
Through a review of existing literature, we categorize previous criteria into two main types. The first type is objective criteria, which are based on information accessible to any observer and are designed to be individual-independent. Examples include goal attainment \cite{nixon1992impact, 10.1145/3449247, telepresence}, agenda completeness \cite{WebRTC}, and the presence of a clear purpose and structure \cite{ComFeel}. A special case is research built on the ``Winter Survival Task'' \cite{rebuttal1, rebuttal2, rebuttal6}, where meeting performance is quantified by the absolute deviation of group rankings from expert rankings, yielding a highly objective yet task-specific score for each meeting. These approaches aim for consistent evaluations across different raters. The second type is subjective criteria, which are assessed by asking participants to rate based on their personal feelings and context, and therefore, individual-dependent. Examples include achieving personal work goals \cite{Not_Another_Meeting, Leach2009Perceived} and decision satisfaction \cite{nixon1992impact}. Such criteria cannot be evaluated solely from the observable context of the meeting. 
Current evaluation approaches fail to capture the dynamic nature of collaborative sessions and often use context-specific criteria which are not universally applicable.
In addition, a number of studies have explored leveraging multimodal cues to automatically predict meeting outcomes \cite{rebuttal3, rebuttal4, rebuttal5, rebuttal7, rebuttal8}, though they still rely on the aforementioned criteria and produce a single holistic score.

\textbf{LLM-as-a-Judge.} 
The paradigm of using LLMs as evaluators, or ``LLM-as-a-Judge'' \cite{initial_llm_judge}, has emerged as a prominent and scalable approach for approximating human assessment. By leveraging in-context learning, they can perform evaluations that demonstrate strong alignment with human judgments \cite{gu2025surveyllmasajudge}. This methodology has been successfully applied across diverse applications, from Natural Language Generation evaluation \cite{gao-etal-2025-llm, chen-etal-2023-exploring-use, liu-etal-2023-g} - such as dialogue system evaluation and text summarization assessment - to specialized domains such as finance, law, education, and science \cite{10.1145/3630106.3659048, 10.1145/3706468.3706507, zhou-etal-2024-llm}. To our knowledge, this is the first work to apply the LLMs-as-a-judge paradigm for meeting effectiveness evaluation.

\section{Meeting Effectiveness Criteria} 

In this section, we identify the most fundamental aspect of evaluating meeting effectiveness: the evaluation criteria. The scope of what constitutes a meeting is exceptionally broad. \citet{Meeting_analysis} define a meeting as ``a focused interaction of cognitive attention, planned or chance, where people agree to come together for a \textit{common purpose}, whether at the same time and the same place, or at different times in different places.'' This definition encompasses a wide range of interactions, from formal board meetings to casual hallway conversations, and includes gatherings aimed at decision-making as well as those intended to improve interpersonal relationships. Our goal is to establish a universal evaluation criterion applicable to any type of meeting.

Based on the two categories of criteria introduced in Section \ref{subsec:mee} and considering the difficulty in accessing participants' personal dispositions, we adopt an objective evaluation criterion. 
Specifically, we propose a criterion based on two factors: meeting objectives achievement and time utilization. This can be expressed in the following formula: Meeting Effectiveness =  Objectives Achievement / Time Cost. This formulation aligns with the fundamental definition of ``efficiency,'' which is the ratio of outcome to cost. In the context of a meeting, the primary outcome is the achievement of the ``common purpose'' as defined earlier, while the principal cost is the time invested by participants. Prior studies relying on human evaluation rarely modeled time explicitly. Instead, time was implicitly factored into metrics like ``satisfaction'' or ``effectiveness,'' relying on the annotator's intuition to account for duration. However, for objective automatic evaluation, explicit timing is essential.

Crucially, the ``objectives achievement'' is a holistic measure of collective progress toward all objectives, implicitly weighing their relative importance, rather than assessing each one individually.
And we define the ``meeting objectives'' as the goals synthesized from the meeting's content upon its conclusion, rather than those planned beforehand. This accommodates the natural evolution of meetings, where new urgent issues can arise and shift the focus. In such cases, effectiveness is evaluated against these emergent objectives, not as a deviation from the original objectives.

\begin{figure*}[t]
  \centering
  \begin{subfigure}[b]{0.25\linewidth}
    \includegraphics[width=\linewidth]{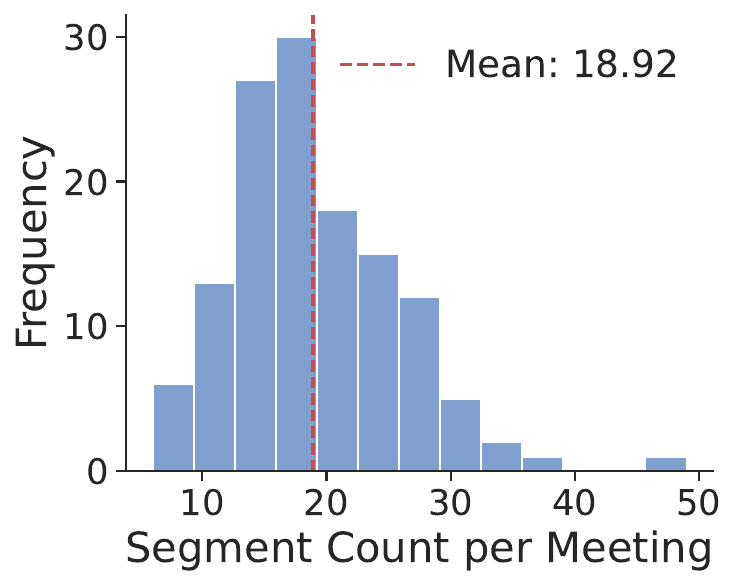}
    \vspace{-7mm}
    \caption{}
    \label{fig:seg_count_dist}
  \end{subfigure}
  \hfill
  \begin{subfigure}[b]{0.25\linewidth}
    \includegraphics[width=\linewidth]{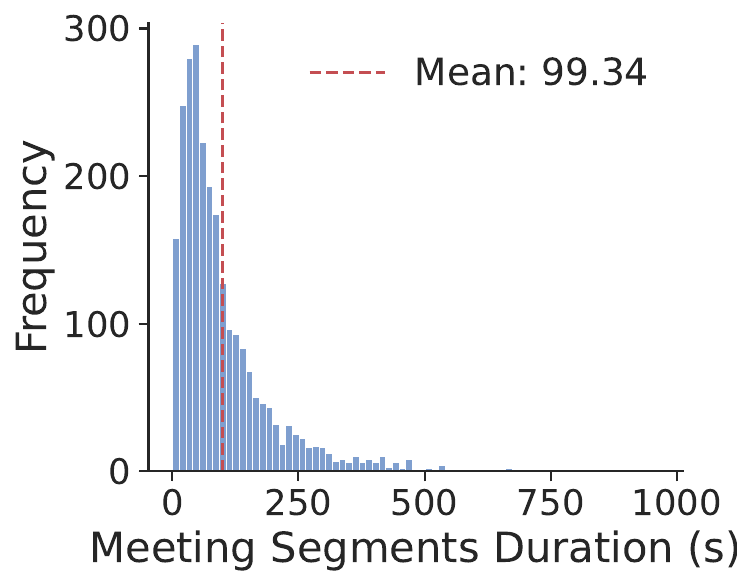}
    \vspace{-7mm}
    \caption{}
    \label{fig:seg_duration_dist}
  \end{subfigure}
  \hfill
  \begin{subfigure}[b]{0.23\linewidth}
    \includegraphics[width=\linewidth]{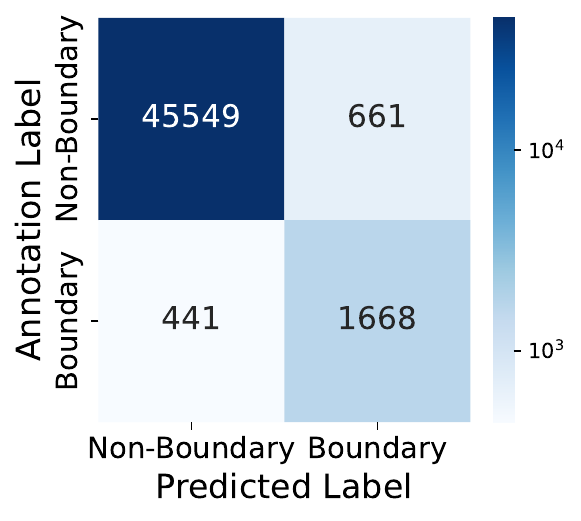}
    \vspace{-7mm}
    \caption{}
    \label{fig:confusion_matrix}
  \end{subfigure}
  \hfill
  \begin{subfigure}[b]{0.25\linewidth}
    \includegraphics[width=\linewidth]{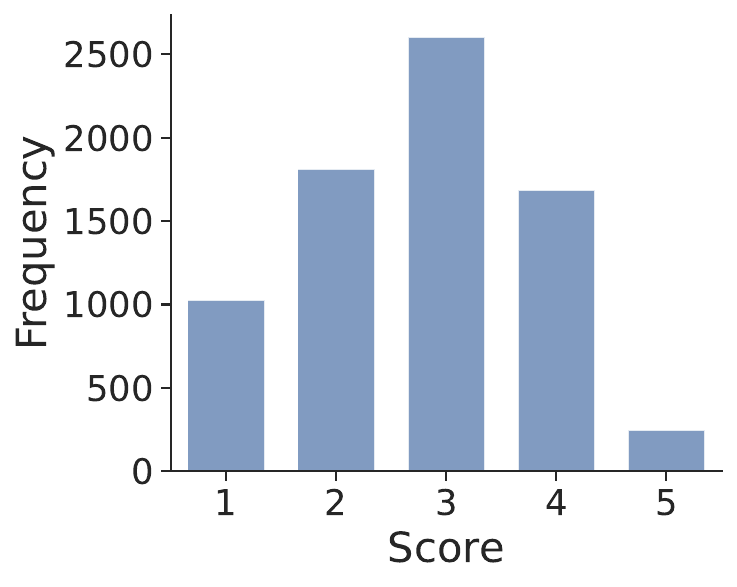}
    \vspace{-7mm}
    \caption{}
    \label{fig:seg_score_dist}
  \end{subfigure}
  
  \caption{Statistics of the AMI-ME dataset. (a) Distribution of segment count per meeting. (b) Distribution of segment duration. (c) Confusion matrix for topic segmentation, viewed as a binary classification task of identifying boundaries between adjacent utterances. (d) Distribution of segment effectiveness scores.}
  \label{fig:dataset}
\end{figure*}

\section{Temporal Fine-grained Evaluation}

Existing approaches to meeting effectiveness evaluation typically assign a single holistic score to an entire meeting. In contrast, we propose a temporal fine-grained evaluation method, which involves dividing the meeting into distinct segments and evaluating the effectiveness of each one. 
This method offers several advantages. First, it enables more detailed analysis. A meeting may have periods of high efficiency interspersed with periods of low efficiency. A single score would obscure this nuance, leading to a significant loss of valuable information. Second, fine-grained evaluation enhances rating accuracy, as shorter focused segments are easier for annotators to assess. Finally, this method substantially increases the volume of available data. Research in meeting science has historically been constrained by data scarcity. Meetings are time-consuming to conduct, and their content often contains private information, making it difficult to collect and publicly release real-world data. If each meeting yields only a single data point, conducting large-scale quantitative analysis becomes challenging. By segmenting each meeting, we can significantly increase the number of data points, thereby facilitating more robust statistical analysis.

We typically segment meetings based on discussion topics. A crucial consideration in this approach is the segmentation granularity. An overly coarse-grained segmentation presents two primary drawbacks: (1) it reduces scoring reliability, as a long segment may contain sub-parts of varying effectiveness, making a single rating difficult and imprecise; and (2) it results in the loss of information specific to each sub-part.
Conversely, an excessively fine-grained segmentation can lead to inaccurate evaluations due to a lack of context. However, this drawback can be mitigated by providing the necessary context to raters. Therefore, we adopt a fine-grained segmentation where each segment represents a minimal topical unit with no clear internal boundaries that would permit further division.
The overall effectiveness of a meeting can then be calculated as the duration-weighted average of its segment-level scores, as derived in Appendix \ref{sec:overall_score}.

\vspace{-1mm}
\section{Dataset Construction}
\vspace{-1mm}

In this section, we introduce the construction of our meta-evaluation dataset for meeting effectiveness, the AMI-ME dataset. The process involves selecting a suitable corpus, performing reference-based topic segmentation, and conducting human annotation of segment effectiveness.

\begin{figure*}[t]
  \centering
  \includegraphics[width=0.7\linewidth]{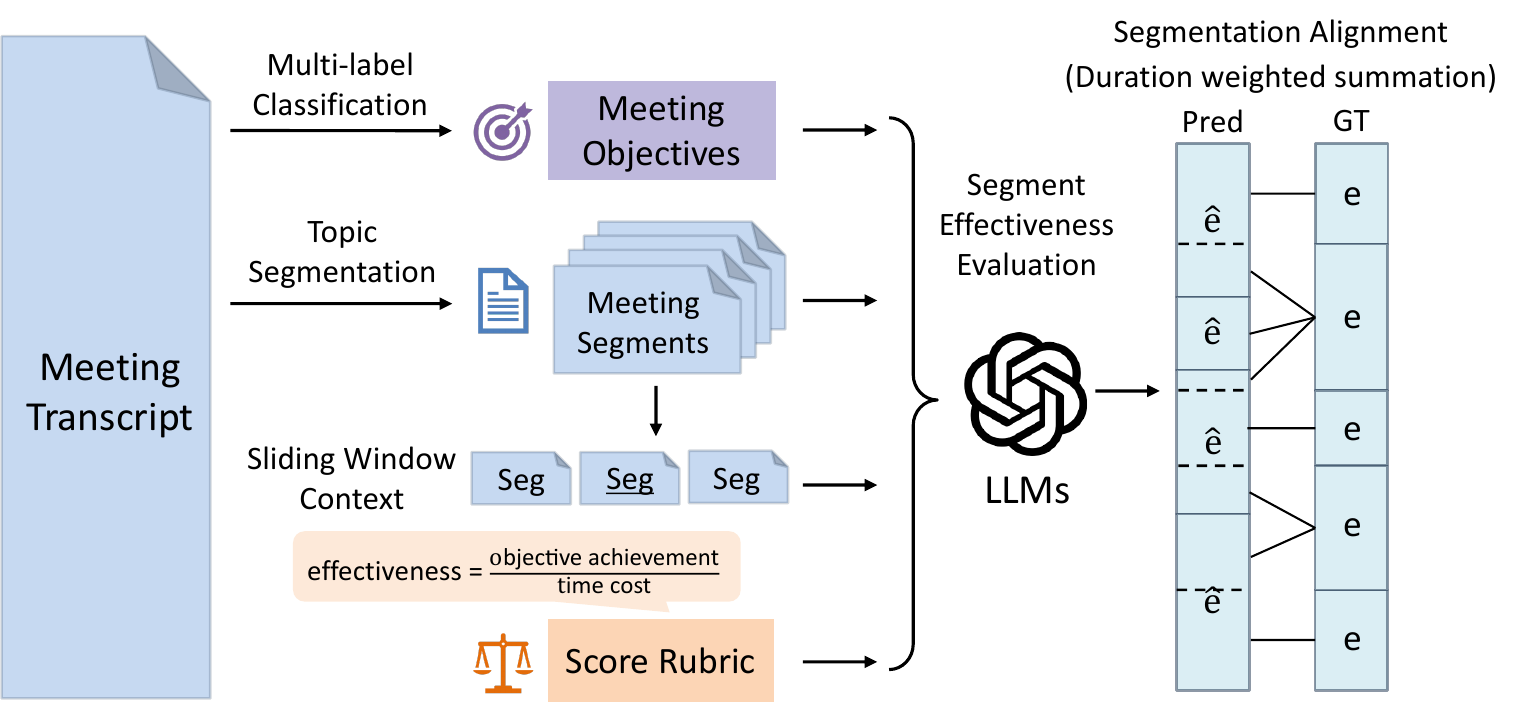}
  \vspace{3mm}
  \caption{The automatic evaluation framework.}
  \label{fig:framework}
\end{figure*}

\subsection{Reference-based Topic Segmentation}

We built our dataset upon the AMI Corpus \cite{AMI}, which mainly consists of highly realistic simulated business meetings and is therefore accessible to annotators without requiring specialized domain knowledge. From the AMI Corpus, we selected 130 meetings for our dataset. These meetings typically involve four participants and last 31 minutes on average. Further details are provided in Appendix \ref{sec:AMI}.

Although the AMI Corpus provides topic segmentation annotations, they are not suitable for our fine-grained evaluation. The existing annotations are hierarchical, with top-level topics that are sometimes further divided into subtopics. However, we identified two issues with these annotations: (1) subtopic segments are often discontinuous, leaving some utterances unassigned, and (2) the granularity of some top-level segments is too coarse, lacking necessary subtopic divisions.

To address these limitations, we developed a reference-based topic segmentation method. This method uses an LLM to produce continuous and more fine-grained segmentations, with the original AMI annotations serving as a reference in the prompt. We retained all existing top-level segmentations from the AMI Corpus and refined the subtopic segmentation. The prompt also instructed the LLM to generate a topic label and a brief description for each subtopic segment. 

After a comparative analysis (see Appendix \ref{sec:ref_top_seg}), we selected Gemini-2.5-Pro for this task. The distributions of the resulting segment count and duration are shown in Figures \ref{fig:dataset}(a) and \ref{fig:dataset}(b). Our manual review confirmed that the model made few critical errors, such as placing a boundary in the middle of a cohesive discussion. Its main weakness was occasionally producing a segmentation that was coarser than the fine-grained level our specific task requires. 
As shown in Figure \ref{fig:dataset}(c), the resulting segmentation incorporated 1,668 of the 2,109 original boundaries and introduced 661 new ones, omitting many original boundaries to ensure continuity.

\subsection{Human Annotations for Effectiveness}
After segmentation, we collected human annotations for segment effectiveness through a rigorous quality control process. 
Given the complexity of the task and the quality differences between crowd-sourced and expert annotations reported in previous work \cite{gillick-liu-2010-non, bhandari-etal-2020-evaluating}, we hired a professional annotation company, conducted a training program for all potential annotators, and selected the top six performers from a qualification test to ensure high aptitude for this task. The six annotators were divided into two groups of three and each group was assigned a distinct set of meetings.

The annotation interface provided comprehensive context, including the evaluation criteria, the segmented transcript, a set of manually pre-defined meeting objectives, and extensive human-written or LLM-generated auxiliary information to aid meeting understanding. 
The pre-defined meeting objectives were manually designed for each type of AMI meeting.
In the interface, each utterance was displayed with timestamps, enabling annotators to consider the time cost of a segment when assessing its effectiveness. 
For each segment, the annotators identified the objectives it served and provided an effectiveness score on a 5-point scale.
To mitigate potential rating skew associated with purely numerical scales \cite{rating_skew}, we used positive-skewed worded (more positive-than negative-valenced verbal labels) options as the rating options. 
More details are available in Appendix \ref{sec:interface}.

\subsection{Dataset Analysis}
The resulting AMI-ME dataset contains 2,459 segments from 130 meetings. Each segment has three independent effectiveness scores. The distribution of effectiveness scores is shown in Figure \ref{fig:dataset}(d). In addition to effectiveness scores, the dataset also includes multi-label annotations identifying the meeting objectives addressed by each segment.

To assess inter-annotator agreement, we calculated the Intraclass Correlation Coefficient (ICC) using a two-way random effects model based on absolute agreement for the average of $k$ raters (ICC($2$,$k$)) \cite{Koo2016AGO}. The first annotation group of three annotators annotated 63 meetings and achieved an ICC of $0.8769$, while the second group of three annotators annotated 67 meetings and achieved an ICC of $0.8202$. These values indicate a ``Good'' level of reliability.

\section{Automatic Evaluation Framework}

We designed a novel framework that leverages LLMs for the automatic evaluation of meeting effectiveness, as illustrated in Figure \ref{fig:framework}. The framework first identifies the objectives for the entire meeting via multi-label classification, then performs topic segmentation, and finally evaluates each segment's effectiveness based on the overall objectives. Considering the discrepancy in predicted segmentation and ground truth segmentation, we also implement a segmentation alignment procedure to support accurate meta-evaluation.

\subsection{Meeting Objective Classification}
We begin by identifying the objectives of each meeting. To ensure the objectives are both generalizable and controllable, we model this problem as a multi-label classification task. We adopted a predefined set of 19 meeting objectives proposed by \citet{telepresence}. Validated through a comprehensive literature review and validated via in-depth interviews that achieved theoretical saturation, this taxonomy captures the comprehensive space of meeting purposes. Furthermore, being rooted in fundamental social psychology theories \cite{McGrath}, these objectives represent universal dimensions of human group interaction—such as conflict resolution and idea generation—ensuring their generalizability beyond commercial settings to the diverse scenarios. The detailed objectives are provided in Appendix \ref{sec:prompts}.
Given the highly free-form nature of meetings, setting a manual upper limit on the number of objectives encourages the model to focus on the most salient goals. Specifically, for the AMI Corpus, we constrain the model to select a maximum of three objectives per meeting through a manual inspection of various meeting types within the dataset. We acknowledge that for particularly complex meetings outside this dataset a higher limit can be required.

Furthermore, our preliminary explorations showed that a naive prompting approach over many categories was prone to positional bias, where the model preferentially selects labels that appear early in the list. To mitigate this, we designed a structured prompting strategy that first identifies a broad set of all potentially relevant objectives, and then analyzes textual evidence to refine the selections to the most central ones. This method prevents the model from defaulting to early options and ensures a more robust classification. The prompt is available in Appendix \ref{sec:prompts}.

\subsection{Segment Effectiveness Evaluation} 
To evaluate segment effectiveness, we employ a method inspired by G-Eval \cite{liu-etal-2023-g}, which generates evaluation steps with Chain-of-Thought (CoT) and uses a form-filling paradigm. For non-reasoning LLMs, we use the token probability weighted summation of the output score as the final score. For reasoning LLMs, we sample five times and use the average score as the final score. In this way, the final score is a continuous score instead of a discrete one, and it can better capture the subtle difference between the effectiveness of segments.

The input to the model consists of several components: a detailed score rubric defining different levels of effectiveness, the meeting objectives identified in the previous step, the transcript of the target segment, and its surrounding context segments. 
The score rubric shown in Appendix \ref{sec:prompts} defines effectiveness based on how well a segment contributes to the overall meeting objectives while efficiently utilizing time.
Crucially, we provide the full set of meeting objectives defined for the entire meeting, not just those relevant to the segment, to encourage a holistic evaluation that implicitly accounts for the relative importance of different objectives. 
To provide sufficient context for evaluation, we employ a sliding window method, presenting the target segment along with a fixed number of preceding and succeeding segments. This ensures the model has sufficient surrounding discourse to understand the segment's purpose and trajectory within the broader conversation.
Based on these inputs, the model assesses how effectively the segment contributed to the overall meeting objectives while efficiently utilizing time, and assigns a corresponding effectiveness score.

\begin{table*}[t]
\small
\begin{center}
\setlength{\tabcolsep}{4.5pt} 
\begin{tabular}{@{}l rrr rrr rrr@{}}
\toprule
 & \multicolumn{2}{c}{\textbf{Scenario Meetings}} & \multicolumn{2}{c}{\textbf{Non-Scenario Meetings}} & \multicolumn{2}{c}{\textbf{All Meetings}} \\ 
\cmidrule(lr){2-3} \cmidrule(lr){4-5} \cmidrule(l){6-7} 
\textbf{Model} & \textbf{Spearman ($\boldsymbol{\rho}$)} & \textbf{Kendall ($\boldsymbol{\tau}$)} & \textbf{Spearman ($\boldsymbol{\rho}$)} & \textbf{Kendall ($\boldsymbol{\tau}$)} & \textbf{Spearman ($\boldsymbol{\rho}$)} & \textbf{Kendall ($\boldsymbol{\tau}$)} \\ \midrule
Llama3.3-70B-Instruct & \underline{0.6558} & \textbf{0.5381} & 0.2411\rlap{$^{\dag}$} & 0.2031 & 0.6072\rlap{$^{\dag\ddag}$} & \textbf{0.4854} \\
DeepSeek-R1-70B & 0.6298\rlap{$^{\dag\ddag}$} & 0.4905 & \underline{0.3852} & \underline{0.3012} & 0.6132\rlap{$^{\dag\ddag}$} & 0.4663 \\
Qwen3-32B (reasoning) & 0.6278\rlap{$^{\dag\ddag}$} & 0.4805 & 0.3193\rlap{$^{\dag}$} & 0.2456 & 0.6113\rlap{$^{\dag\ddag}$} & 0.4578 \\
Qwen3-32B (non-reasoning) & \textbf{0.6594} & \underline{0.4939} & \textbf{0.4564} & \textbf{0.3347} & \textbf{0.6445} & \underline{0.4803} \\ \midrule
GPT-4o & 0.6518 & 0.4923 & 0.2873\rlap{$^{\dag}$} & 0.2074\rlap{$^{\dag}$} & \underline{0.6341} & 0.4756 \\
Gemini-2.5-Flash & 0.5793\rlap{$^{\dag\ddag}$} & 0.4262\rlap{$^{\dag\ddag}$} & 0.1845\rlap{$^{\dag}$} & 0.1289\rlap{$^{\dag}$} & 0.5624\rlap{$^{\dag\ddag}$} & 0.4122\rlap{$^{\dag\ddag}$} \\ \bottomrule
\end{tabular}
\end{center}
\caption{Meta-evaluation of effectiveness scoring using ground truth inputs. ${\dag}$ and ${\ddag}$ indicates a significant difference compared to Qwen3-32B (non-reasoning) and GPT-4o at p < 0.05, respectively.}
\label{tab:GT_inputs}
\end{table*}

\subsection{Segmentation Alignment}

A critical challenge in meta-evaluating the segment-level effectiveness score arises from the discrepancy between model-generated and ground-truth segmentation boundaries. Because the number and duration of segments often differ, a direct one-to-one comparison is not feasible. 
To address this, we implement a segmentation alignment procedure that maps the predicted scores onto the ground truth segments, ensuring a fair and accurate comparison. 

Formally, let the ground truth segments be $\{[t_i, t_{i+1}]\}_{i=0}^{n-1}$, with a corresponding set of effectiveness scores $\{e_{t_i}^{t_{i+1}}\}_{i=0}^{n-1}$. Similarly, let the predicted segments be $\{[\hat{t}_j, \hat{t}_{j+1}]\}_{j=0}^{\hat{n}-1}$, with corresponding predicted scores $\{\hat{e}_{\hat{t}_j}^{\hat{t}_{j+1}}\}_{j=0}^{\hat{n}-1}$.

For each $i$-th ground truth segment, we compute a single aligned prediction score $\hat{e}_{t_i}^{t_{i+1}}$ derived from all prediction segments that have a temporal overlap with it. This aligned score is a weighted average of the scores from the overlapping prediction segments, where the weight is determined by the duration of the overlap:
\begin{equation}
\hat{e}_{t_i}^{t_{i+1}} = \frac{\sum_{j=0}^{\hat{n}-1} \hat{e}_{\hat{t}_j}^{\hat{t}_{j+1}} \cdot \Delta_{i,j}}{\sum_{j=0}^{\hat{n}-1} \Delta_{i,j}}
\end{equation}
where $\Delta_{i,j}$ represents the duration of the temporal overlap between the $i$-th ground truth segment and the $j$-th predicted segment.

In this process, the contribution of each overlapping predicted segment to the final aligned score is proportional to the duration of the overlap.
This process yields a list of aligned prediction scores, $\{\hat{e}_{t_i}^{t_{i+1}}\}_{i=0}^{n-1}$, that is parallel to the ground truth scores, $\{e_{t_i}^{t_{i+1}}\}_{i=0}^{n-1}$, enabling the direct computation of correlation metrics.

\vspace{-1mm}
\section{Experiments}
\vspace{-1mm}

\subsection{Settings}
To establish a comprehensive benchmark, we selected a diverse suite of leading open-source and proprietary LLMs, detailed in Appendix \ref{sec:settings}. This selection ensures a robust and wide-ranging comparison of the capabilities of current state-of-the-art models.
For all LLMs, we adopted their recommended hyperparameter settings to ensure optimal performance.

For topic segmentation metrics, we adopted Pk \cite{Pk} and WindowDiff (Wd) \cite{Wd}. For meeting objective classification, we adopted Hamming Loss and Micro-F1 score. Metric details are available in Appendix \ref{sec:settings}. We evaluated the correlation coefficient between predicted and ground truth effectiveness scores using segment-level Spearman ($\rho$) and Kendall ($\tau$), where the ground truth score for each segment is the mean of its annotation scores.

\begin{table*}[]
\small
\begin{center}
\begin{tabular}{@{}lrrrr@{}}
\toprule
\textbf{} & \multicolumn{2}{c}{\textbf{Measured Correlation}} & \multicolumn{2}{c}{\textbf{Theoretical Upper Bound}} \\ \cmidrule(lr){2-3} \cmidrule(l){4-5}
Model & \textbf{Spearman ($\boldsymbol{\rho}$)} & \textbf{Kendall ($\boldsymbol{\tau}$)} & \textbf{Spearman ($\boldsymbol{\rho}$)} & \textbf{Kendall ($\boldsymbol{\tau}$)} \\ \midrule
Qwen3-32B (non-reasoning) & 0.6445 & 0.4803 & 1.0000 & 1.0000 \\
+ From Segmentation & 0.2256\rlap{$^{\dag}$} & 0.1604\rlap{$^{\dag}$} & 0.6417 & 0.5095 \\
+ From Speech & 0.2180\rlap{$^{\dag}$} & 0.1547\rlap{$^{\dag}$} & 0.6640 & 0.5291 \\ \midrule
GPT-4o & 0.6341 & 0.4756 & 1.0000 & 1.0000 \\
+ From Segmentation & 0.2360\rlap{$^{\dag}$} & 0.1664\rlap{$^{\dag}$} & 0.6828 & 0.5513 \\
+ From Speech & 0.2006\rlap{$^{\dag}$} & 0.1430\rlap{$^{\dag}$} & 0.6725 & 0.5397 \\ \bottomrule
\end{tabular}
\end{center}
\caption{Meta-evaluation and upper bound estimation of effectiveness scoring using upstream inputs. ${\dag}$ indicates a significant difference compared to the ground truth inputs at p < 0.05.}
\label{tab:from_upstream}
\end{table*}

\begin{table}
\small
\begin{center}
\begin{tabular}{@{}lrr@{}}
\toprule
\textbf{model} & \textbf{Wd $\downarrow$}& \textbf{Pk $\downarrow$} \\ \midrule
Absence & 0.4176 & 0.4176 \\
BERTSeg & 0.4139 & 0.4058 \\
SumSeg & 0.4069 & 0.4002 \\ \midrule
Llama3.3-70B-Instruct & 0.4108 & 0.3727\rlap{$^{\dag}$} \\
Qwen3-32B (non-reasoning) & 0.3795\rlap{$^{\dag}$} & \underline{0.3402}\rlap{$^{\dag}$} \\
Qwen3-32B (reasoning) & \underline{0.3713}\rlap{$^{\dag}$} & 0.3418\rlap{$^{\dag}$} \\
DeepSeek-R1-70B & \textbf{0.3658}\rlap{$^{\dag}$} & \textbf{0.3325}\rlap{$^{\dag}$} \\ \midrule
GPT-4o & 0.3980 & 0.3548\rlap{$^{\dag}$} \\
Gemini-2.5-Flash (non-reasoning) & 0.4486 & 0.3576\rlap{$^{\dag}$} \\ \bottomrule
\end{tabular}
\end{center}
\caption{Experiments of topic segmentation. ${\dag}$ indicates a
significant difference compared to Absence, BERTSeg, and SumSeg at p < 0.05, respectively.}
\label{tab:seg}
\end{table}

\subsection{Benchmarking LLMs}
\label{sec:gt_inputs}

To isolate and benchmark the core effectiveness evaluation capabilities of different LLMs, we conducted experiments on all meetings under ideal conditions. We provided all models with identical ground truth topic segmentations and meeting objectives, thereby removing the influence of upstream task performance. We set context window size as 1. Ablation study of context window size and meeting objectives is available in Appendix \ref{sec:ablation}.

As shown in the ``All Meetings'' columns of Table \ref{tab:GT_inputs}, the results demonstrate a high degree of consistency across most models, with Qwen3-32B (non-reasoning) achieving the highest Spearman correlation. 
Gemini-2.5-Flash is a notable exception, exhibiting markedly lower correlation, which we attribute to its tendency to overuse the lowest possible score.
While Llama3.3 achieves the highest Kendall score, this result is influenced by a highly concentrated token probability distribution for its scores, which function more as discrete than continuous values. We measured the average variance of the token probability distributions across all segments and found that Llama3.3 had variances of only $0.06$. In contrast, all other models had variances exceeding $0.28$. Such a concentrated distribution leads to many ties in the scores. This results in a higher Kendall correlation without necessarily reflecting a superior evaluation capability \cite{liu-etal-2023-g}.
Further analysis of segment duration and score is shown in Appendix \ref{sec:score_dur} and an analysis of the inter-LLM scoring consistency is provided in Appendix \ref{sec:inter_LLM}.

\subsection{Analysis across Meeting Types}
\label{sec:non_scenario}
To investigate the framework's performance on different meeting types and the framework's generalizability, we isolated the scenario and non-scenario meetings from the AMI Corpus for separate analysis. Unlike the scenario business meetings, the non-scenario meetings are long unscripted discussions on topics of film selection and office relocation. They lack a workflow or a designated role, making both topic segmentation and effectiveness evaluation significantly more difficult. In particular, topic segmentation for these unstructured interactions proves challenging even for human annotators.

As shown in Table \ref{tab:GT_inputs}, although the correlation scores for non-scenario meetings are lower than those for scenario meetings due to the increased complexity of the source material, the relative performance trends across models remained consistent with the experiments on all meetings. This consistency validates the framework's generalizability, suggesting that it effectively captures the universal aspects of meeting effectiveness across distinct interaction structures, rather than being suitable only for specific business scenarios.

\subsection{Impact of Automatic Topic Segmentation}

We investigated the framework's performance under model-predicted segmentations. We first benchmarked various unsupervised topic segmentation models, including a baseline with no boundaries (``Absence''), BERTSeg \cite{BERTSeg}, and SumSeg \cite{artemiev-etal-2024-leveraging}, alongside various LLMs. For BERTSeg and SumSeg, we search for hyperparameters with the official implementation using Optuna \cite{akiba2019optuna}. For LLM-based topic segmentation, the core context of the prompt is available in Appendix \ref{sec:prompts}. As shown in Table \ref{tab:seg}, LLMs significantly outperform non-LLM baselines. The lower performance of Gemini-2.5-Flash is due to its tendency to over-segment, producing an average of $28.7$ segments per meeting compared to the ground truth of $18.9$. For subsequent experiments, each LLM's effectiveness evaluation was conducted based on its own segmentation output.

We proceeded following experiments with the best performing open-source and proprietary models from Section \ref{sec:gt_inputs}: Qwen3 (non-reasoning) and GPT-4o. As shown in the ``Measured Correlation'' column of Table \ref{tab:from_upstream}, the correlation scores based on predicted segmentations (+ From Segmentation) decrease substantially compared to those using ground truth segmentation. 

To isolate the performance loss caused by segmentation mismatches, we calculated an approximate correlation upper bound given the predicted segmentations. This was done by aligning the ground truth effectiveness scores to the predicted segmentation, treating these aligned scores as a new prediction, and then re-aligning them back to the ground truth segmentation for correlation calculation. The resulting values are reported in the ``Theoretical Upper Bound'' column of Table \ref{tab:from_upstream}. Notably, these upper bounds (e.g., Spearman of $0.6417$ for Qwen3) are significantly lower than the perfect correlation of $1.0$. This demonstrates that the structural discrepancy between predicted and ground truth segmentations inherently penalizes the evaluation metric. The current alignment score essentially functions as a combined metric that enforces requirements for both fine-grained segmentation and scoring accuracy, which is precisely what our proposed paradigm necessitates. The upper bound estimation quantifies the relative impact of the two factors, providing valuable guidance for future improvements. A detailed analysis of how segmentation granularity influences the upper bound is available in Appendix \ref{sec:propagation}.

\subsection{End-to-End System Performance}
Finally, we evaluated the performance of a full end-to-end system, starting from raw speech audio. Our pipeline begins with voice activity detection and speaker diarization using pyannote \cite{pyannote}, followed by Automatic Speech Recognition (ASR) on the resulting active speech intervals with Whisper-Large-V3 \cite{whisper}. The transcribed text is then processed through topic segmentation, objective classification, and finally, effectiveness evaluation.

We first evaluated the performance of the upstream speech processing components. For speaker diarization, we calculated the Diarization Error Rate with a collar of $0.25$ seconds, which resulted in $24.03\%$. For ASR, we obtained a Word Error Rate of $31.95\%$ after JiWER's standard text normalization \cite{Morris2004FromWA} and removing punctuation. These performance levels are comparable to those reported in previous studies \cite{pyannote, imai-etal-2025-evaluating}, indicating a solid performance of our speech processing part.

As shown in Table \ref{tab:from_upstream} (+ From Speech), the end-to-end system's correlation scores are comparable to those starting from ground-truth transcripts (+ From Segmentation). This suggests that the effectiveness evaluation component is relatively robust to speech transcription errors, which we attribute to the inherent resilience of LLMs to noisy text.
\vspace{-0.5mm}
\section{Conclusion}
\vspace{-1mm}
In this paper, we introduced a new paradigm for meeting effectiveness evaluation. To support this paradigm, we constructed a meta-evaluation dataset and proposed an LLM-based framework for automatic evaluation. Our experiments establish a comprehensive benchmark, demonstrating that LLMs can achieve strong correlation with human judgments. By establishing baselines for various models and end-to-end systems, our work provides a solid foundation for future research in meeting analysis and intelligent multi-party dialogue agents.

\section*{Limitations}
Our study has three primary limitations.
First, a comprehensive meeting evaluation should not only consider effectiveness, but also cover the well-being of participants, such as psychological safety and inclusiveness. We believe incorporating participant well-being is a valuable direction for future work.
The second limitation is about the linearity assumption in effectiveness scoring. Our definition of effectiveness and the subsequent segmentation alignment method implicitly assume that the human-annotated scores have a linear relationship with the degree of objective achievement. However, it is challenging for human annotators to maintain a perfectly linear scale in their judgments. If the underlying rating scale is non-linear, the validity of the weighted averaging could be compromised. Future work could investigate mapping the scores to a latent linear domain before alignment and then mapping the result back to the original scale to achieve higher accuracy. 
Third, our experiments are conducted on the AMI Corpus, which was selected for annotation requirements and the high degree of realism. However, they are simulated business meetings that may not fully capture the higher stakes, implicit power hierarchies, organizational politics, and culturally contingent social norms of real-world interactions. 

\section*{Ethical Considerations}
This work adheres to ethical guidelines for data collection and distribution. The dataset, AMI-ME, is derived from the publicly available AMI Corpus. In line with the original corpus, we will release the dataset under the Creative Commons Attribution 4.0 license agreement (CC BY 4.0), promoting open access and reproducibility. The annotations collected do not contain any personally identifiable information about the annotators, ensuring their privacy and anonymity.
Furthermore, we acknowledge the potential risks of performative behavior associated with the deployment of automated meeting effectiveness evaluations. If participants are aware that their interactions are being scored, they may artificially alter their behavior. This awareness could inadvertently incentivize rushing through agendas or dominating the conversation to appear productive, thereby undermining genuine collaboration and psychological safety. Future real-world applications of this framework must carefully consider these dynamics to ensure that evaluation metrics support rather than distort healthy organizational communication.

\section*{Acknowledgements}
This work was supported by JST SPRING (Grant Number JPMJSP2110) and JSPS (Grant Number JP23K28144).

\bibliography{custom}

\appendix
\section{The Derivation of Overall Meeting Effectiveness}
\label{sec:overall_score}
Formally, let $o_{a}^{b}$ represent the total objective achievement over a given time interval $[a,b]$. The effectiveness $e_{a}^{b}$ is then the average objective achievement rate within that interval, defined as:
\begin{equation}
 e_{a}^{b}= \frac{o_{a}^{b}}{b-a}
\end{equation}

For a meeting spanning the interval $[0,T]$, we partition this duration into $n$ contiguous segments, denoted by $\{[t_i, t_{i+1}]\}_{i=0}^{n-1}$, where $t_0=0$ and $t_n=T$. 
The overall effectiveness $e_0^T$ can then be derived from the effectiveness of these individual segments:
\begin{align*}
e_0^T   &= \frac{o_0^T}{T} \\
        &= \frac{\sum_{i=0}^{n-1} o_{t_i}^{t_{i+1}}}{T} \\
        &= \sum_{i=0}^{n-1} \left( \frac{o_{t_i}^{t_{i+1}}}{t_{i+1} - t_i} \cdot \frac{t_{i+1} - t_i}{T} \right) \\
        &= \sum_{i=0}^{k-1} e_{t_i}^{t_{i+1}} \cdot \frac{t_{i+1} - t_i}{T}
\end{align*}


This confirms that the overall meeting effectiveness is the weighted average of its segment scores, where each segment's weight is its proportional duration.

\begin{figure*}[t]
  \centering
  \includegraphics[width=\linewidth]{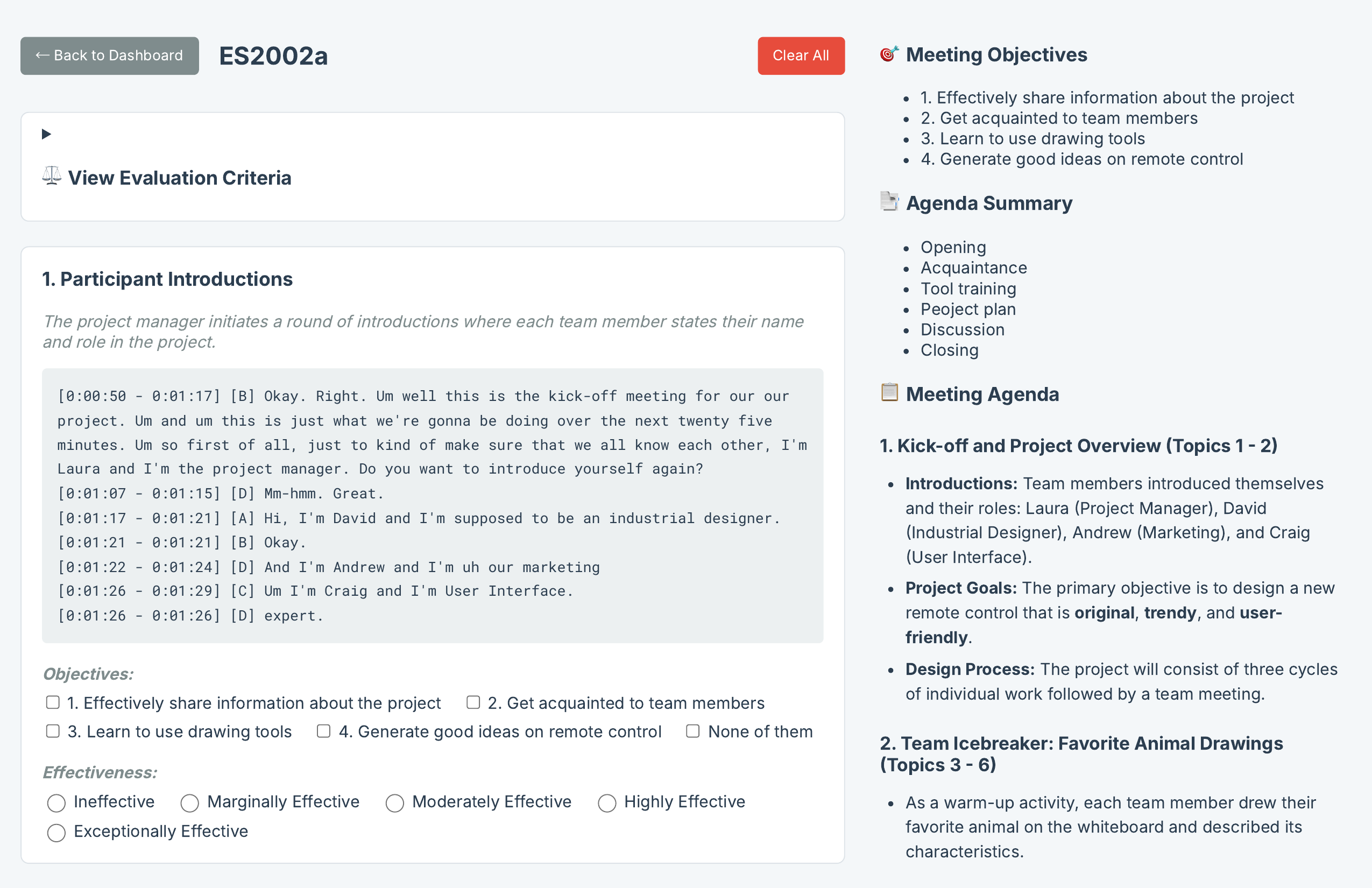}
  \caption{The annotation interface.}
  \label{fig:interface}
\end{figure*}

\section{Additional Details for Dataset Construction}
\subsection{The AMI Corpus}
\label{sec:AMI}

The collection of authentic meeting data is particularly difficult due to the time-intensive nature of meetings and the privacy concerns associated with their content. Consequently, the number of publicly available meeting corpora is limited. Existing datasets can be broadly categorized by their domain, including research group meetings \cite{ICSI, AMI}, political and parliamentary sessions \cite{hu-etal-2023-meetingbank}, and business meetings \cite{AMI}. For our task, which requires human annotators to comprehend the meeting content fully, corpora from specialized domains like research or politics present a significant challenge due to the extensive background knowledge required. Therefore, we chose the AMI Corpus \cite{AMI}, which is centered around business scenarios.

The AMI Corpus is a multimodal dataset comprising 100 hours of meeting recordings. It is enriched with a variety of high-quality annotations, such as manually produced orthographic transcriptions and topic segmentations.

The meetings in the corpus are divided into two types: scenario meetings and non-scenario meetings. In scenario meetings, participants role-play as employees of an electronics company tasked with designing a new television remote control. They are given a brief to develop a prototype over a series of four meetings. Because the participants are not actual experts in product design but are acting out a role, the discussions require relatively few domain-specific background knowledge, making them accessible to annotators. Non-scenario meetings include a variety of topics. The majority are natural research meetings, while a few non-research meetings cover topics such as selecting films for a fictitious movie club, a fictional office relocation discussion, and planning a postgraduate workshop.

Although AMI-scenario meetings are simulated business meetings, the specific workflow is not strictly controlled or scripted, and the overall interaction remains consistent with natural meetings \cite{AMI}. In addition, the AMI scenario meetings are internally divided into four different types, effectively covering diverse meeting categories. For instance, the first meeting type includes ice-breaking activities and tool familiarization, whereas the fourth type involves performance reviews.

From the AMI Corpus, we selected meetings with existing topic segmentations and filtered for completeness, resulting in 130 meetings. This set consists of 126 scenario (business) meetings, 2 film selection meetings, and 2 office relocation meetings. 

\subsection{Model Selection for Reference-based Topic Segmentation}
\label{sec:ref_top_seg}
We evaluated two models for reference-based topic segmentation: Qwen3-32B (reasoning) \cite{qwen3} and Gemini-2.5-Pro \cite{gemini2.5}. A comparative analysis was conducted on five randomly selected meetings. Taking Qwen3's output as a baseline, we identified 24 variations (merges, splits, or boundary shifts) in Gemini-2.5-Pro's segmentation. A review of these variations showed that Gemini-2.5-Pro's output was superior in 14 cases, Qwen3's was better in 7 cases, and they were of comparable quality in 3 cases.

\subsection{The Detailed Annotation Protocol}
\label{sec:interface}

We contracted a professional annotation company called Anosupo to perform the data annotation. The annotators are native English speakers from the Philippines. We paid at a rate of $71.5$ JPY (approximately $0.48$ USD) per segment. We have received the necessary permissions for the public release of this annotated dataset.

The annotation interface, shown in Figure \ref{fig:interface}, was designed to provide annotators with all necessary context and extensive auxiliary information for accurate evaluation. For each segment, annotators were required to answer two questions: (1) A multiple-choice question to identify which meeting objectives the segment serves, with an additional option for ``None of them.''  (2) A single-choice question to rate the segment's effectiveness. Crucially, the holistic assessment required annotators to consider both the progress made on corresponding objectives and the relative importance of that objective to the meeting’s overall objectives.

A central component of the context was the set of predefined meeting objectives. 
The AMI Corpus we used consists of six meeting types (four scenario-based and two non-scenario-based). As meetings within each type exhibit a high degree of consistency in their agendas and objectives, we manually summarized a set of 2-4 objectives for each type based on the actual meeting content. While this summarization was informed by a classification of 19 meeting objectives \cite{telepresence}, the resulting objectives are distinct from this general set. The 19 objectives are broad in scope, whereas our manually summarized objectives are specifically tailored to the AMI Corpus. An example of the four objectives for one meeting type is shown in Figure \ref{fig:interface}. 

To further enhance the annotators' understanding of the meetings, we provided extensive auxiliary information. This included: (1) An annotation guide detailing the AMI Corpus and the interface functionalities. (2) A set of annotation examples as anchors, which provided sample ratings for various segments along with detailed justifications. (3) For each meeting, a handwritten agenda summary corresponding to its meeting type for a high-level overview, and a more detailed LLM-generated agenda that introduces top-level segments with corresponding subtopic range and key achievements. (4) The LLM-generated topic and description for each individual segment. We explicitly informed the annotators that the LLM-generated content was intended as supplementary aids and should be treated as potentially containing errors.

Furthermore, because the annotation mainly covered four distinct types of AMI scenario meetings, we enforced annotators to annotate meetings by type (e.g., annotating all Type A meetings before moving to Type B). This allowed them to develop an intuitive grasp of the reasonable duration for segments within each specific meeting category.

\section{Prompts}
\label{sec:prompts}
\lstdefinestyle{promptstyle}{
  backgroundcolor=\color{black!5},   
  basicstyle=\ttfamily\small,      
  breaklines=true,                 
  rulecolor=\color{black},         
  captionpos=b,                    
  keywordstyle=\color{blue},       
  morekeywords={Human, AI},         
  xleftmargin=0pt,
}

The score rubric used in the automatic evaluation framework is shown as follows:
\begin{tcolorbox}[
  colback=gray!5,  
  colframe=gray!50,
  title=Prompt,
  fonttitle=\bfseries,
  breakable,        
  boxrule=0.5pt     
]
\small\ttfamily
Effectiveness (1-5): the effectiveness of the meeting segmentation in terms of how effectively it contributed to the overall meeting objectives while efficiently utilizing time.
Formally, Effectiveness = (Objective Achievement)/(Time Cost)
The overall meeting objectives are \{\{OVERALL\_MEETING\_OBJECTIVES\}\}\\

Score 1: Ineffective Segment

- Segment had little or no relevance to the meeting objectives

- Time was poorly utilized with excessive tangents, repetition, or discussions that could have been handled elsewhere

- Participants gained very little value relative to the time invested\\

Score 2: Marginally Effective Segment

- Segment had some connection to the meeting objectives but with limited concrete progress

- Time usage showed clear inefficiencies (unfocused discussion, unclear direction, excessive details)

- Value delivered was noticeably low compared to time spent\\

Score 3: Moderately Effective Segment

- Segment had a clear connection to the meeting objectives with some measurable progress

- Time was reasonably managed with typical pacing and standard level of focus

- The value gained appropriately matched the time invested\\

Score 4: Highly Effective Segment

- Segment made significant progress toward the meeting objectives with clear outcomes

- Time was well utilized with focused discussion and few unnecessary diversions

- The segment delivered good value relative to the time invested\\

Score 5: Exceptionally Effective Segment

- Segment was critical to achieving the meeting objectives with decisive progress

- Time usage was highly efficient, with calibrated discussion depth and focus

- The segment delivered outstanding value for the time invested
\end{tcolorbox}

The meeting objective classification labels \cite{telepresence} are shown as follows:
\begin{tcolorbox}[
  colback=gray!5,   
  colframe=gray!50,
  title=Prompt, 
  fonttitle=\bfseries,
  breakable,       
  boxrule=0.5pt     
]
\small\ttfamily
1. Exchange/share opinions or views on a topic or issue

2. Make a decision

3. Give or receive orders

4. Find a solution to a problem that has arisen

5. Generate ideas on products, projects or initiatives

6. Generate buy-in or consensus on an idea

7. Resolve conflicts and disagreements within a group

8. Build trust and relationships with one or more individuals

9. Maintain relationships with one or more other people and stay in touch

10. Negotiate or bargain on a deal or contract

11. Routine exchange of information

12. Non-routine exchange of information

13. Communicate positive or negative feelings or emotions on a topic or issue

14. Show personal concern about or interest in a particular issue or situation

15. Assert and/or reinforce your authority, status, position to your team or others

16. Give or receive feedback

17. Assemble a team and/or motivate teamwork on a project

18. Clarify a concept, issue or idea

19. Exchange confidential, private or sensitive information
\end{tcolorbox}

The core context of the three-step meeting objective classification prompt is shown as follows:
\begin{tcolorbox}[
  colback=gray!5,   
  colframe=gray!50,
  title=Prompt, 
  fonttitle=\bfseries,
  breakable,       
  boxrule=0.5pt    
]
\small\ttfamily
Three-Round Selection Process:

Round 1 - Identify potentially relevant objectives with their original ID numbers (1-19)

Round 2 - Detailed Analysis: Examine evidence for each candidate objective, eliminate those with minimal support.

Round 3 - Final Selection: From remaining objectives, select up to 3 PRIMARY objectives with strongest evidence.

\end{tcolorbox}

\begin{figure*}[t]
  \centering
  \begin{subfigure}[b]{0.45\linewidth}
    \includegraphics[width=\linewidth]{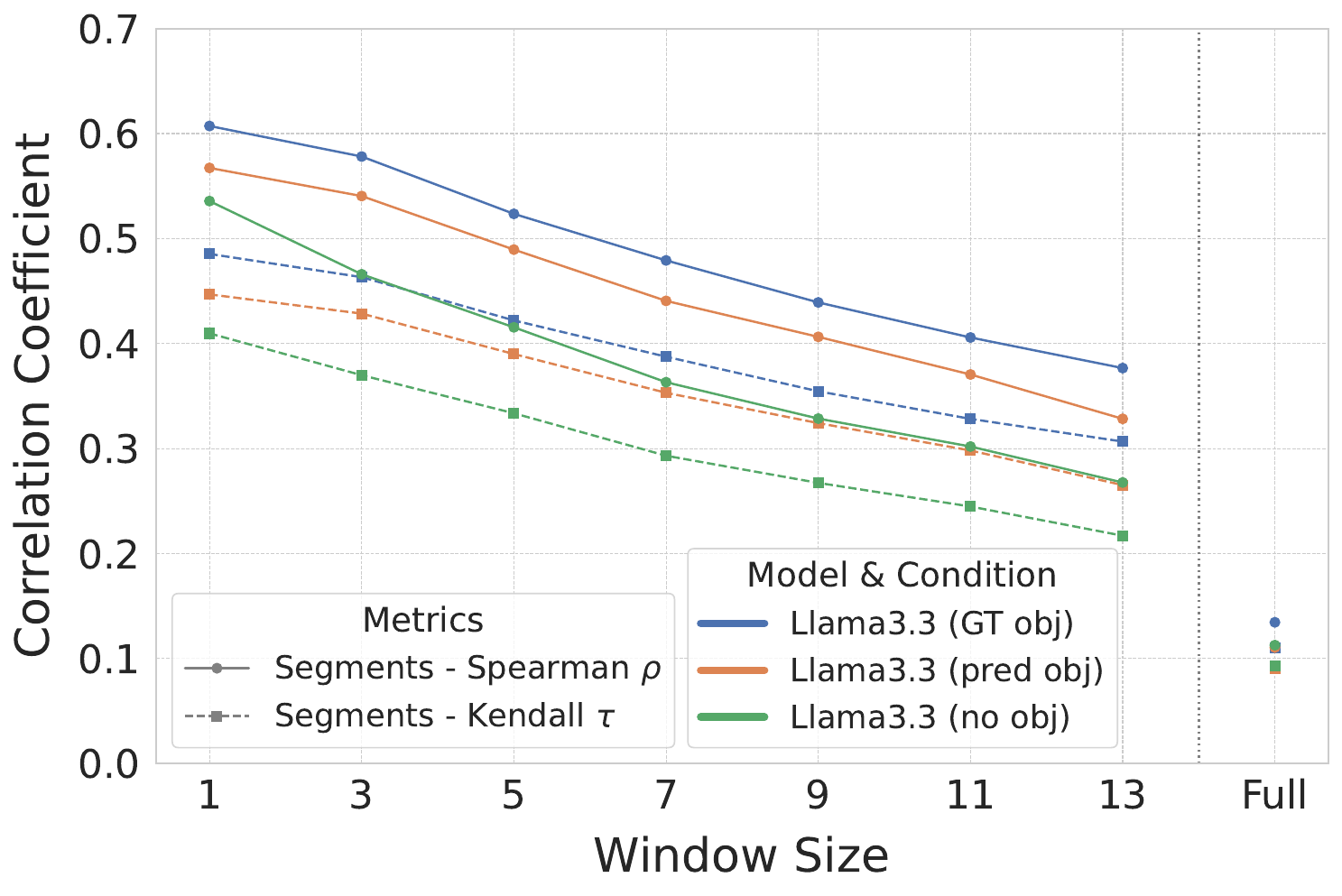}
    \vspace{-7mm}
    \caption{}
    \label{fig:llama_ablation} 
  \end{subfigure}
  \qquad 
  \begin{subfigure}[b]{0.45\linewidth}
    \includegraphics[width=\linewidth]{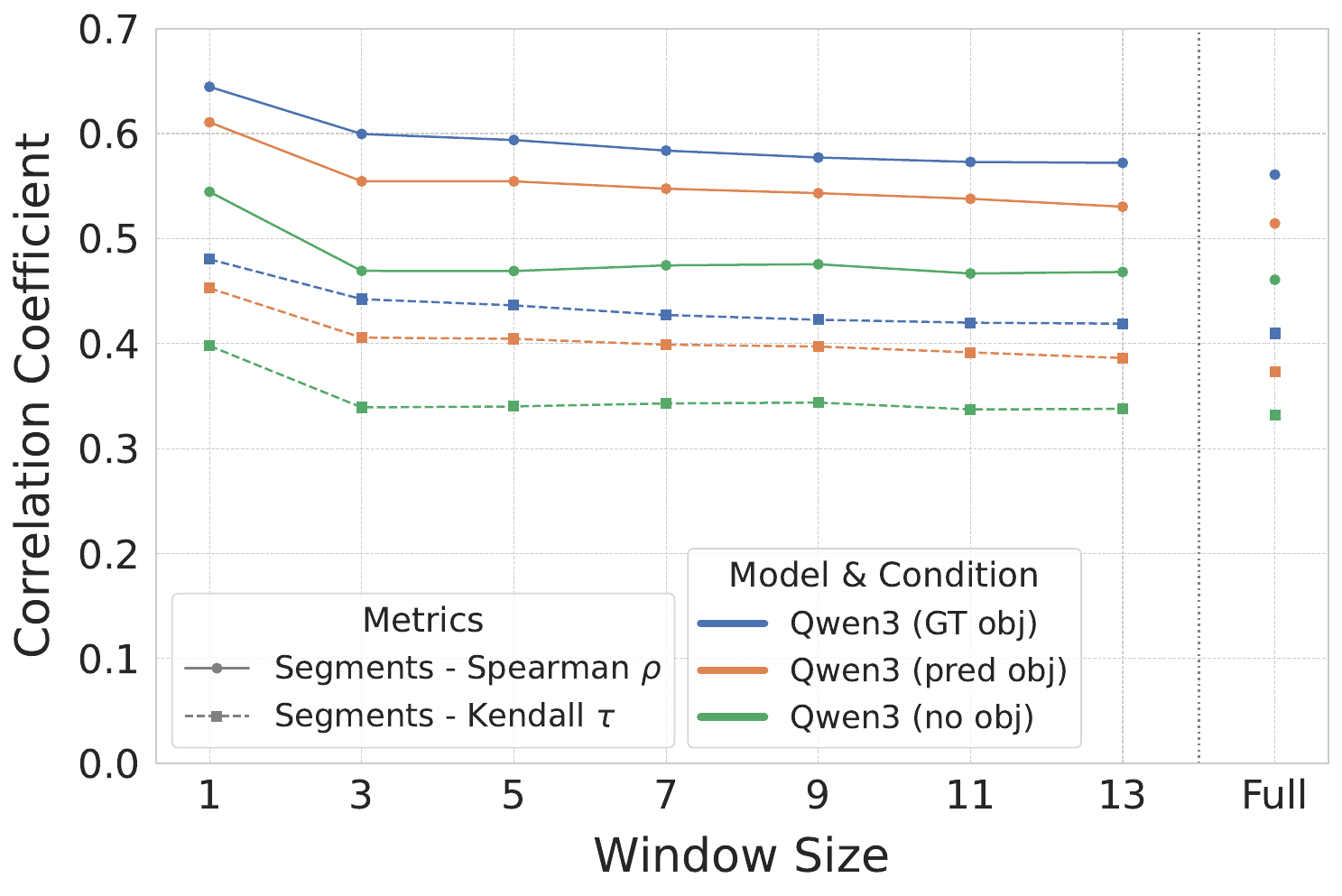}
    \vspace{-7mm}
    \caption{}
    \label{fig:qwen_ablation} 
  \end{subfigure}
  \caption{Ablation studies of the context window size and the meeting objectives. (a) Experiments on Llama3.3-70B-Instruct. (b) Experiments on Qwen3-32B (non-reasoning). The ``full'' window size refers to using the entire transcript as context.}
  \label{fig:ablation_study} 
\end{figure*}

\begin{table*}[t]
\small
\begin{center}
\setlength{\tabcolsep}{4.5pt} 
\begin{tabular}{@{}l rrr rrr rrr@{}}
\toprule
 & \multicolumn{2}{c}{\textbf{Scenario Meetings}} & \multicolumn{2}{c}{\textbf{Non-Scenario Meetings}} & \multicolumn{2}{c}{\textbf{All Meetings}} \\ 
\cmidrule(lr){2-3} \cmidrule(lr){4-5} \cmidrule(l){6-7} 
\textbf{Model} & \textbf{Spearman ($\boldsymbol{\rho}$)} & \textbf{Kendall ($\boldsymbol{\tau}$)} & \textbf{Spearman ($\boldsymbol{\rho}$)} & \textbf{Kendall ($\boldsymbol{\tau}$)} & \textbf{Spearman ($\boldsymbol{\rho}$)} & \textbf{Kendall ($\boldsymbol{\tau}$)} \\ \midrule
Llama3.3-70B-Instruct & 0.6198 & 0.5062 & 0.3827 & 0.3219 & 0.5781 & 0.4633 \\
DeepSeek-R1-70B & 0.6021 & 0.4682 & 0.1959 & 0.1430 & 0.5820 & 0.4409 \\
Qwen3-32B (reasoning) & 0.5999 & 0.4588 & 0.2783 & 0.2046 & 0.5831 & 0.4356 \\
Qwen3-32B (non-reasoning) & 0.6350 & 0.4708 & 0.3326 & 0.2433 & 0.5996 & 0.4423 \\ \midrule
GPT-4o & 0.5757 & 0.4277 & 0.2189 & 0.1586 & 0.5618 & 0.4153 \\
Gemini-2.5-Flash & 0.5427 & 0.3991 & 0.2442 & 0.1772 & 0.5260 & 0.3855 \\ \bottomrule
\end{tabular}
\end{center}
\caption{Meta-evaluation of effectiveness scoring using ground truth inputs with context window size of 3.}
\label{tab:GT_inputs_winsize_3}
\end{table*}

The core context of the topic segmentation prompt is shown as follows:
\begin{tcolorbox}[
  colback=gray!5,  
  colframe=gray!50,
  title=Prompt,
  fonttitle=\bfseries,
  breakable,       
  boxrule=0.5pt   
]
\small\ttfamily 
Please perform **fine-grained topic segmentation** on the meeting transcript.\\

Instructions:

1. Divide the transcript into distinct segments based on topic changes. Ensure each segment represents a coherent topic discussion with clear boundaries for optimal topic segmentation.\\

2. Make the segmentation as fine-grained as possible, identifying even subtle topic shifts, while maintaining topic coherence within each segment.\\

3. For each segment, provide:

    - `start\_id`: The ID of the first utterance of the segment.
    
    - `end\_id`: The ID of the last utterance of the segment.
    
    - `topic`: A concise phrase describing the main topic.
    
    - `description`: A one-sentence summary of the segment content.\\
    
4. Critical Check for Completeness and Continuity:

    - **No Gaps**: The `start\_id` ID of any segment (except the first) must immediately follow the `end\_id` ID of the preceding segment. For example, if segment N ends at ID 15, segment N+1 must start at ID 16.
    
    - **Full Coverage**: All utterances from the first utterance ID provided in the transcript to the very last utterance ID MUST be included in a segment.
    
    - **Final Utterance**: The `end\_id` ID of the very last segment **must** be the ID of the last utterance in the entire `Meeting transcripts`. The last utterance ID in the provided transcript is `\{\{LAST\_UTTERANCE\_ID\_PLACEHOLDER\}\}`. \\

Format your response in a structured JSON array as specified below. Ensure the JSON is valid.
...
\end{tcolorbox}

\section{Experiment Settings}
\label{sec:settings}
We selected a diverse suite of LLMs for experiments. The open-source models evaluated are Llama3.3-70B-Instruct \cite{llama3}, DeepSeek-R1-70B \cite{deepseekr1}, and Qwen3-32B \cite{qwen3}. For proprietary models, we included GPT-4o \cite{gpt4o} and Gemini-2.5-Flash \cite{gemini2.5}. 

All experiments involving open-source models were conducted on a server equipped with four NVIDIA A6000 GPUs. To ensure efficient inference for the open-source models, we utilized the vLLM library \cite{kwon2023efficient}.

For topic segmentation metrics, we adopted Pk \cite{Pk} and WindowDiff (Wd) \cite{Wd}. 
Both metrics operate by moving a sliding window across the document or transcripts. Pk calculates a penalty by assessing whether the endpoints of the window are correctly placed in the same or different segments relative to the ground truth. In contrast, Wd penalizes differences in the number of segment boundaries that fall within the window when comparing the predicted segmentation to the ground truth.

For meeting objective classification metrics, we adopted Hamming Loss and Micro-F1 score. The ground truth labels were derived from the manually summarized objectives created during the human annotation phase. As these objectives do not directly align with the predefined 19-category classification schema, we established a mapping from each ground truth objective to a set of relevant categories. For instance, the ground truth objective ``Generate good ideas on remote control'' could correspond to multiple labels in the schema, such as ``Generate ideas on products, projects, or initiatives,'' ``Exchange/share opinions or views on a topic or issue,'' and ``Find a solution to a problem that has arisen.'' To account for this one-to-many relationship, we framed the evaluation as a bipartite graph matching problem, assessing the alignment between the predicted labels and the ground truth label sets, from which the final scores are computed.

\begin{figure*}[t]
  \centering
  \begin{subfigure}[b]{0.48\linewidth}
    \includegraphics[width=\linewidth]{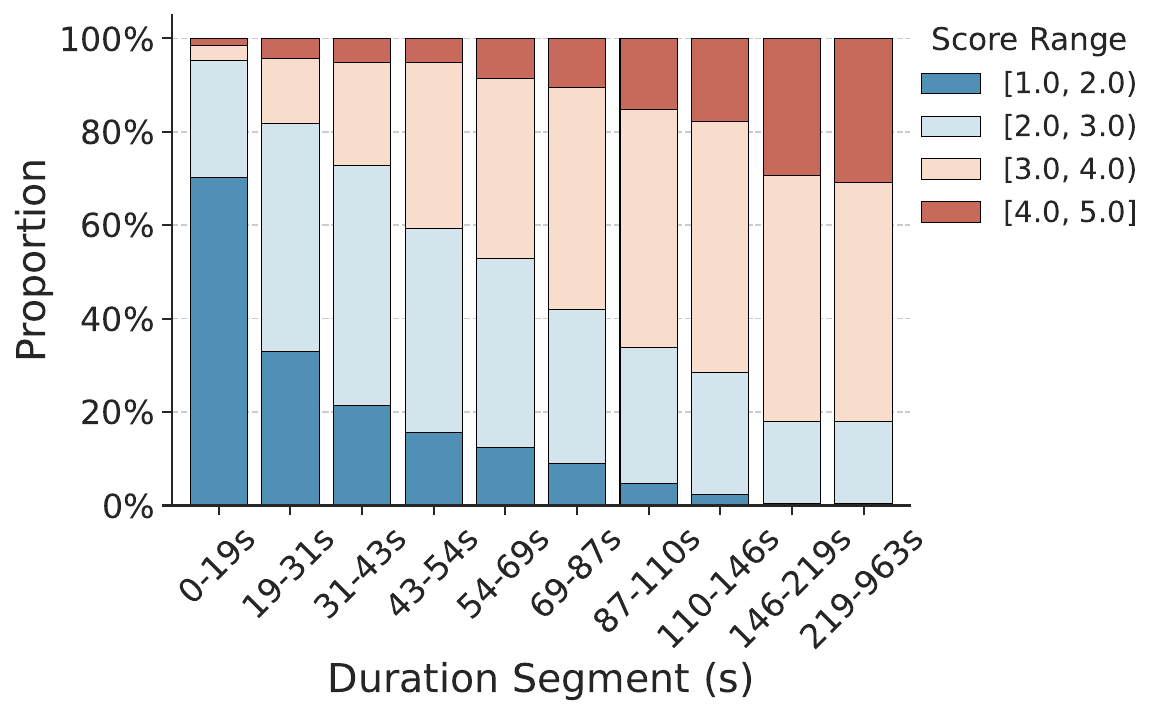}
    \vspace{-7mm}
    \caption{}
    \label{fig:score_dur_gt}
  \end{subfigure}
  \hfill
  \begin{subfigure}[b]{0.48\linewidth}
    \includegraphics[width=\linewidth]{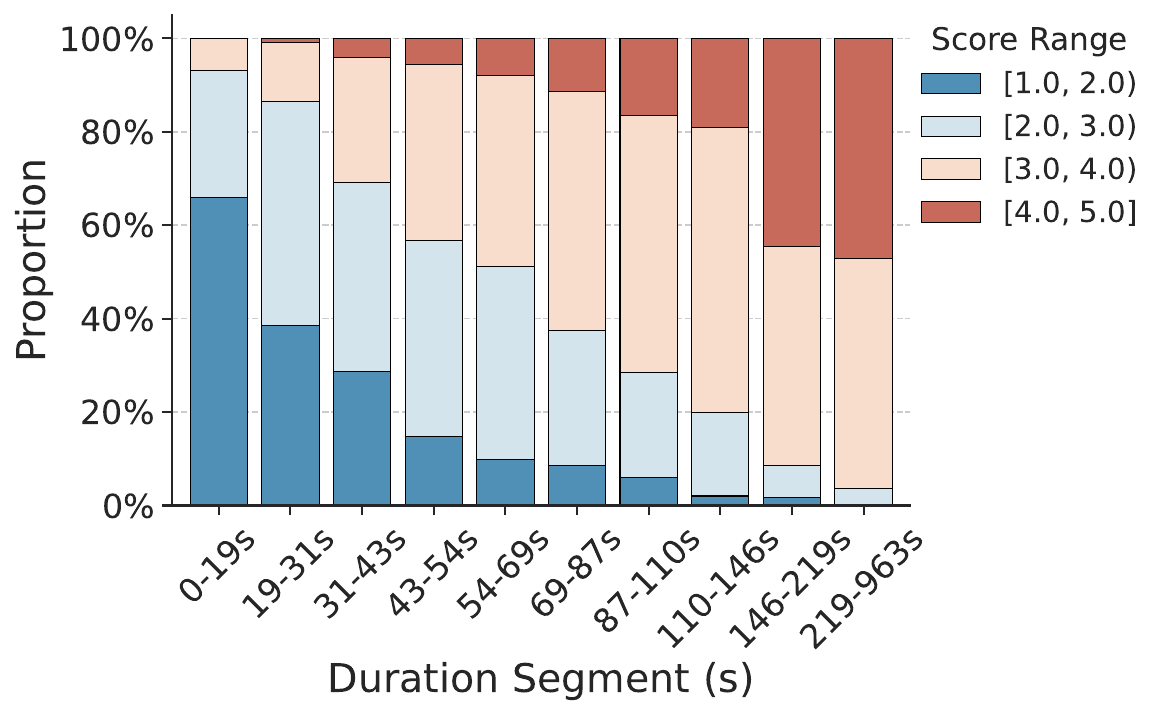}
    \vspace{-7mm}
    \caption{}
    \label{fig:score_dur_pred}
  \end{subfigure}
  \hfill
  \caption{Relationship between segment scores and duration. (a) Human annotation. (b) Prediction of Qwen3-32B (non-reasoning) with context window size = 1.}
  \label{fig:score_dur}
\end{figure*}

\section{Ablation Studies}
\label{sec:ablation}
We conducted ablation studies to investigate the impact of two critical components of the automatic evaluation framework: the context window size and the meeting objectives. These experiments used the Qwen3-32B (non-reasoning) and Llama3.3-70B-Instruct models and the ground truth topic segmentation. For meeting objectives, we compared three conditions: (1) providing ground truth objectives (GT obj), (2) providing objectives predicted by the respective LLM (pred obj), and (3) providing no objectives (no obj).

We first report the performance of meeting objective classification. The Qwen3 model achieved a Hamming Loss of $0.08$ and a Micro-F1 of $0.81$, while the Llama3.3 model achieved a Hamming Loss of $0.07$ and a Micro-F1 of $0.86$. These results indicate that both models are proficient at identifying the correct meeting objectives.

The ablation results are presented in Figure \ref{fig:ablation_study}. Regarding meeting objectives, providing ground truth objectives yields the best performance, followed by predicted objectives, and then no objectives. This hierarchy demonstrates the importance of well-defined objectives in guiding the effectiveness evaluation.
For context, a window size of 1 was optimal for both models. 
We hypothesize that the self-sufficiency and completeness of the ground truth segments make additional context unnecessary. The steady performance decline for Llama3 may also suggest a limitation in its ability to process longer contexts effectively. 

The results for different models with context window size of 3 are available in Table \ref{tab:GT_inputs_winsize_3}.

\section{Analysis of Segment Score and Duration}
\label{sec:score_dur}
We analyzed the relationship between segment duration and effectiveness scores to verify whether the incorporation of time cost introduces a systematic bias against longer segments. However, as illustrated in Figure \ref{fig:score_dur}, we observe the opposite: longer segments consistently yield higher scores in both ground truth and prediction settings. This finding alleviates concerns regarding length bias, indicating that the criterion does not simply punish duration. Instead, it reflects the nature of the data, where longer segments typically encapsulate substantive high-value discussions, while extremely short segments often correspond to solving equipment issue or transitional exchanges.

\section{Inter-LLM Scoring Consistency}
\label{sec:inter_LLM}
\begin{figure}[t]
  \centering
  \includegraphics[width=0.98\linewidth]{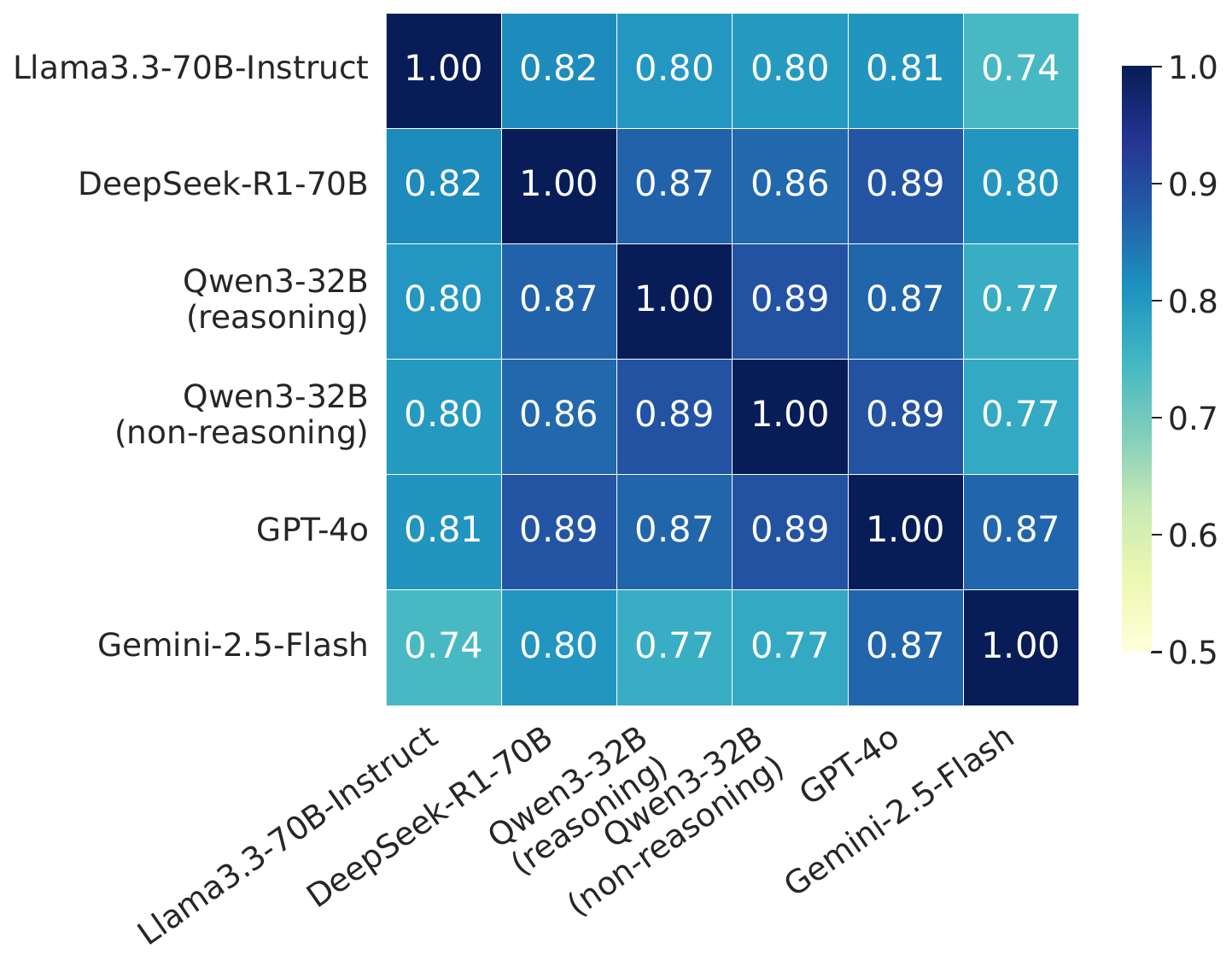}
  \caption{The Spearman correlation coefficient between LLMs with ground truth input on all meetings.}
  \label{fig:inter_LLM}
\end{figure}

To examine the internal consistency among LLMs, we computed pairwise Spearman correlation coefficients between the scores produced by all LLMs using ground truth inputs on the full set of meetings. As shown in Figure \ref{fig:inter_LLM}, inter-LLM consistency is notably higher than human–LLM consistency. In particular, DeepSeek, Qwen3 (both reasoning and non-reasoning), and GPT-4o form a cluster of high mutual consistency.

\begin{figure}[t]
    \centering
    \begin{subfigure}[b]{0.65\linewidth}
        \centering
        \includegraphics[width=\textwidth]{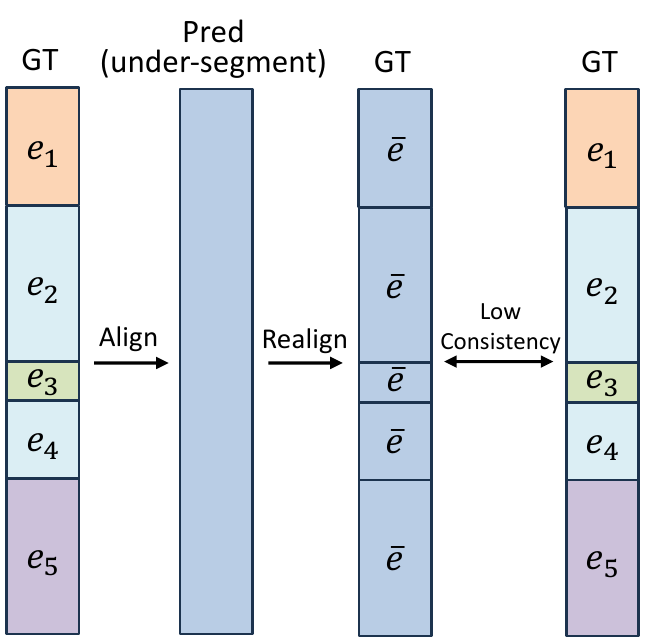}
        \caption{Coarse-grained predicted segmentation yields a low upper bound.}
        \label{fig:single_suba}
    \end{subfigure}
    
    \vspace{0.2cm}
    
    \begin{subfigure}[b]{0.65\linewidth}
        \centering
        \includegraphics[width=\textwidth]{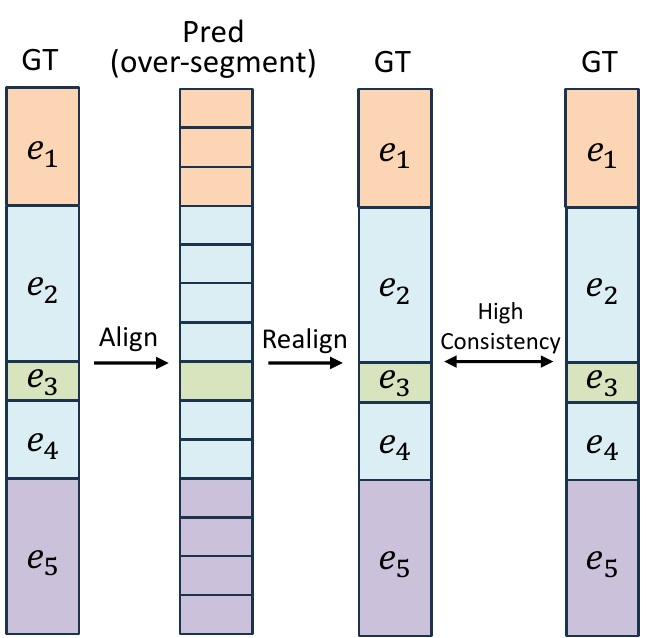}
        \caption{Fine-grained predicted segmentation yields a high upper bound.}
        \label{fig:single_subb}
    \end{subfigure}
    
    \caption{Illustration of how segmentation granularity affects the theoretical upper bound of the alignment-based correlation. Blocks represent segments, and $e$ denote effectiveness scores.}
    \label{fig:propagation}
\end{figure}

\section{Segmentation Error Propagation Analysis}
\label{sec:propagation}
The theoretical upper bound estimation addresses the following question: given the ground truth segmentation with its associated scores and a predicted segmentation, what is the maximum achievable correlation coefficient over all possible score assignments to the predicted segments? Computing the exact upper bound is intractable, as the objective itself involves both the segmentation alignment procedure and the subsequent correlation coefficient computation. The method we employ provides a computationally efficient approximation that yields a loose upper bound, which is less than or equal to the exact upper bound.

The relationship between segmentation granularity and the theoretical upper bound can be understood through two contrasting cases. In general, finer segmentation granularity raises the theoretical upper bound by providing more degrees of freedom for score assignment, though it simultaneously increases the difficulty of accurate scoring for the model. Conversely, coarser granularity or boundary misalignment reduces the theoretical upper bound by constraining the space of achievable score distributions.

To build intuition, consider two extreme cases as shown in Figure \ref{fig:propagation}. First, suppose the predicted segmentation consists of a single monolithic segment spanning the entire meeting (i.e., no segmentation is performed). After alignment, every ground truth segment would be assigned the same effectiveness score, inevitably resulting in zero correlation. Second, suppose the segmentation is extremely fine-grained, such as treating each individual utterance as a separate segment. In this case, the large number of independently assignable scores provides sufficient degrees of freedom such that there exists a score assignment that, after alignment, can perfectly reconstruct the ground truth scores, yielding a correlation of $1.0$.

\end{document}